\documentclass[journal]{IEEEtran}
 
\usepackage{hyperref}
\usepackage{amsmath,amsfonts,amssymb}
\usepackage{verbatim}
\usepackage{algpseudocode}
\usepackage{graphicx}
\usepackage{textcomp}
\usepackage{multirow} 
\usepackage{epstopdf}
\usepackage{color}
\usepackage{algorithm}
\usepackage{threeparttable}
\usepackage[table,xcdraw]{xcolor}
\usepackage{subfigure}
\usepackage{changepage}
\usepackage{bm}
\usepackage{diagbox}
\usepackage{booktabs}
\usepackage{wrapfig}
\usepackage[switch]{lineno}

\newcommand{\mTa}{\mathbf{\Theta}}
\newcommand{\wt}{\mathbf{w}}
\newcommand{\bbC}{\mathbb{C}}
\newcommand{\bht}{\mathbf{h}}
\newcommand{\fd}{\mathrm{d}}
\newcommand{\fr}{\mathrm{r}}
\newcommand{\Gt}{\mathbf{G}}

\begin{document}
\title{AutoOptLib: Tailoring Metaheuristic Optimizers via Automated Algorithm Design}
       
\author{Qi~Zhao,~
        Bai~Yan,~
        Taiwei~Hu,~
        Xianglong~Chen,~
        Qiqi~Duan,~
        Jian~Yang,~
	and~Yuhui~Shi,~\IEEEmembership{Fellow,~IEEE}	

\thanks{This work is supported by the Shenzhen Fundamental Research Program under Grant No. JCYJ20200109141235597, Guangdong Basic and Applied Basic Research Foundation under Grant No. 2021A1515110024, National Natural Science Foundation of China under Grants No. 61761136008, Shenzhen Peacock Plan under Grant No. KQTD2016112514355531, and Program for Guangdong Introducing Innovative and Entrepreneurial Teams under Grant No. 2017ZT07X386. \textit{(Corresponding author: Yuhui Shi)}}
\thanks{Q. Zhao, X. Chen, J. Yang and Y. Shi are with the Department of Computer Science and Engineering, Southern University of Science and Technology, Shenzhen 518055, China (email: zhaoq@sustech.edu.cn; chenxianglong.cs@outlook.com; yangj33@sustech.edu.cn; shiyh@sustech.edu.cn).}
\thanks{B. Yan is with the Research Institute of Trustworthy Autonomous Systems, Southern University of Science and Technology, Shenzhen 518055, China (email: yanb@sustech.edu.cn).}
\thanks{T. Hu is with the Department of Computer Science, Johns Hopkins University, Baltimore, Maryland, USA (email:thu26@jhu.edu).}
\thanks{Q. Duan is with the School of Computer Science and Technology, Harbin Institute of Technology, Shenzhen 518055, China (email:11749325@mail.sustech.edu.cn).}}

\maketitle

\begin{abstract}
Metaheuristics are prominent gradient-free optimizers for solving hard problems that do not meet the rigorous mathematical assumptions of analytical solvers. The canonical manual optimizer design could be laborious, untraceable and error-prone, let alone human experts are not always available. This arises increasing interest and demand in automating the optimizer design process. In response, this paper proposes AutoOptLib, the first platform for accessible automated design of metaheuristic optimizers. AutoOptLib leverages computing resources to conceive, build up, and verify the design choices of the optimizers. It requires much less labor resources and expertise than manual design, democratizing satisfactory metaheuristic optimizers to a much broader range of researchers and practitioners. Furthermore, by fully exploring the design choices with computing resources, AutoOptLib has the potential to surpass human experience, subsequently gaining enhanced performance compared with human problem-solving. To realize the automated design, AutoOptLib provides 1) a rich library of metaheuristic components for continuous, discrete, and permutation problems; 2) a flexible algorithm representation for evolving diverse algorithm structures; 3) different design objectives and techniques for different optimization scenarios; and 4) a graphic user interface for accessibility and practicability. AutoOptLib is fully written in Matlab/Octave; its source code and documentation are available at \url{https://github.com/qz89/AutoOpt} and \url{https://AutoOpt.readthedocs.io/}, respectively.
\end{abstract}

\begin{IEEEkeywords}
Metaheuristic, Optimization, Automated Algorithm Design, Evolutionary Algorithm, Swarm Intelligence Algorithm
\end{IEEEkeywords}

\section{Introduction}
Metaheuristics are stochastic search approaches that integrate gradient-free local improvement with high-level strategies of escaping from local optima \cite{glover2006handbook}. In contrast to analytical methods and problem-specific heuristics, metaheuristics can conduct search on any problem with available solution representation, solution quality evaluation, and a certain notion of locality\footnote{Locality denotes the ability to generate neighboring solutions via a heuristically-informed function of one or more incumbent solutions \cite{swan2022metaheuristics}.} \cite{swan2022metaheuristics}. This has enabled metaheuristic optimizers to be widely recognized for hard problems featured with non-differentiation, multimodality, discretization, and large scale, such as optimizing representations \cite{schmidt2009distilling,weiel2021dynamic}, model structures \cite{stanley2019designing}, and control decisions \cite{birattari2020disentangling}.

Carefully tailoring metaheuristic optimizers to the target problem is necessary to obtain good enough solutions, especially for discrete ones. The tailoring is usually human-made, which suffers from challenges in front of the rapid development of modern science and engineering. First, manual tailoring could be laborious, which may cost the experts days or weeks tailoring the algorithms. Second, manual tailoring may be error-prone due to the high complexity of the target problem and the high degree of freedom in tailoring the algorithms. Third, the manual process is untraceable regarding what motivates certain design decisions, losing insights and principles for future reuse. Lastly, eligible algorithm experts are not always available, especially in practical scenarios.

These challenges motivate automatically designing the optimizers with little human intervention. The automated design leverages computing resources to partly replace human experts to conceive, build up, and verify the design choices of the optimizers to fit for solving the target problem. Human experts can be involved but are not necessary during the design process. Therefore, the automated design could make high-performance algorithms accessible to a much broader range of researchers and practitioners. Moreover, by leveraging computing power to fully explore the design choices, the automated design has the potential to reach or even surpass human experience, thereby gaining enhanced performance compared with human problem-solving.

\textbf{Related work:} Different with automated algorithm selection \cite{2018Automated} and automated parameter configuration \cite{schede2022survey}, in which the former selects algorithms from a portfolio and the latter configures parameters of a given algorithm, automated algorithm design could find either instantiations/variants of existing algorithms or unseen algorithms with diverse structures \cite{stutzle2019automated,2020The,zhao2023survey}. For metaheuristics, the automated design is also related to adaptive operator selection and hyperheuristics that leverage high-level search, learning, and data mining (e.g., exploratory landscape analysis) methods to generate/select low-level heuristics (usually) online by the search trajectory. In contrast, the automated design tailors algorithms offline by a target distribution of problem instances, which is significant for problem-solving scenarios where one can afford a priori computational resources (for design) to subsequently solve many problem instances drawn from the target domain. 

Several platforms have been introduced to promote the accessibility of the automated algorithm design techniques in the metaheuristic community, including irace \cite{lopez2016irace}, Paradiseo \cite{cahon2004paradiseo,dreo2021paradiseo}, ParamILS \cite{hutter2009paramils}, SMAC \cite{hutter2011sequential,lindauer2022smac3}, and Sparkle \cite{van2022sparkle}. Among them, irace provides an R package and proposes the iterative racing mechanism to identify competitive design choices. Paradiseo offers a C++ interface to irace. ParamILS has a Ruby implementation that uses iterative local search to search over the design space. SMAC was written in Python and leverages Bayesian optimization to search for promising design choices. The intensification and capping strategies are involved in SMAC to save evaluations. Sparkle is a Python package that provides easily accessible algorithm configuration by SMAC\footnote{Sparkle also supports automated algorithm selection, which is a different topic from algorithm design.}. 

Most of the platforms stem from hyperparameter configuration and bear restrictions in designing metaheuristic algorithms. First, they mainly specialize in a single module rather than supporting the whole design process. For example, irace emphasises selecting design choices by iterative racing; ParamILS and SMAC specialize in searching design choices and saving algorithm evaluations. Users have to collect design choices and construct the design space themselves, which is nontrivial for practitioners and researchers out of the metaheuristic community. Second, they provide little support for designing algorithms with diverse structures. They manipulate the concatenation of categorical identifiers of algorithm components and numerals of hyperparameters; a predefined algorithm template is required to map the concatenation to its phenotype, limiting the designed algorithms' novelty and diversity. Third, they are stand-alone implementations of certain design techniques rather than uniform software with various accessible techniques.

\textbf{Contribution:} This paper presents AutoOptLib, aiming to eliminate the restrictions. AutoOptLib's contributions include:
\begin{itemize}
    \item \textit{Rich library of design choices.} We provide over 40 metaheuristic algorithm components for continuous, discrete, and permutation problems with/without constraints. Users do not need to write additional code but can directly build their design space through the components.
    \item \textit{Flexibility to designing diverse algorithms.} We manipulate a directed graph representation of the algorithm to be designed; the graph typology evolves during design, enabling discovering algorithms with novel and diverse structures (e.g., unfolded algorithms, algorithms with inner loops, algorithms with recursive operators, and ensemble algorithms with parallel operators) in a single run of design. 
    \item \textit{Various design objectives and techniques.} We cover different types of design objectives, e.g., solution quality, runtime, and anytime performance, and various design techniques, e.g., racing, intensification, and surrogate, for different demands and interests.
    \item \textit{Accessibility to both researchers and practitioners.} We fully write AutoOptLib in MATLAB/Octave to directly interface with abundant MATLAB/Octave simulation environments, which enables usability to researchers from different communities. A Python interface is also provided. Furthermore, we provide a graphic user interface (GUI), allowing users to manage the algorithm design process with simple one-click operations.
    \item \textit{Extensibility.} AutoOptLib follows the open-closed principle \cite{meyer1997object,larman2001protected}. Users are convenient to implement their own design choices and techniques based on the current sources and add the implementations to AutoOptLib by a uniform interface.
\end{itemize}

\textbf{Impact:} Human experts may cost days or weeks to conceive, build up, and verify the optimizers; AutoOptLib saves such labor resources and time costs with today's increasing computational power. Furthermore, through automated algorithm design techniques, AutoOptLib democratizes the efficient and effective use of metaheuristic optimizers. This is significant for researchers and practitioners with complicated optimization problem-solving demands but without the expertise to distinguish and manage suitable optimizers among various choices. In addition, with a uniform collection of related techniques, AutoOptLib would promote research of the automated algorithm design and metaheuristic fields and be a tool in the pursuit of autonomous and general artificial intelligence systems.

The rest of the paper contains: the architecture of AutoOptLib in Section \ref{sec_architecture}, the operating scheme of AutoOptLib in Section \ref{sec_operating}, the running ways of AutoOptLib in Section \ref{sec_run}, additional uses of AutoOptLib in Section \ref{sec_other}, experimental studies in Section \ref{sec_application}, and conclusions in Section \ref{sec_conclusion}.

\section{Architecture} \label{sec_architecture}
\begin{figure*}[htbp] 
	\centering
	\includegraphics[width=0.7\linewidth]{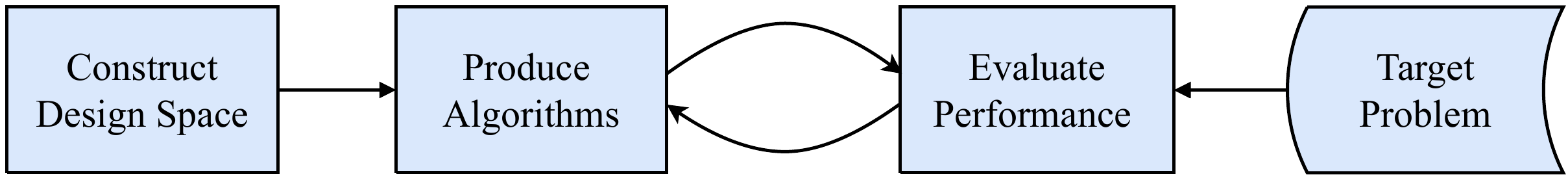}
	\caption{Modules of AutoOptLib.}
	\label{pipeline}
\end{figure*}

AutoOptLib follows a module-based architecture, as shown in Figure \ref{pipeline}. 

\subsection{Construct Design Space} 
AutoOptLib provides over 40 widely-used metaheuristic algorithm components, as given in Table \ref{component}. The components can be used to design optimizers for continuous, discrete, and permutation problems. Users can either employ the default design space with all components for each problem type or define a space according to interest. 

\begin{figure}[t]
\centering
\includegraphics[width=0.9\linewidth]{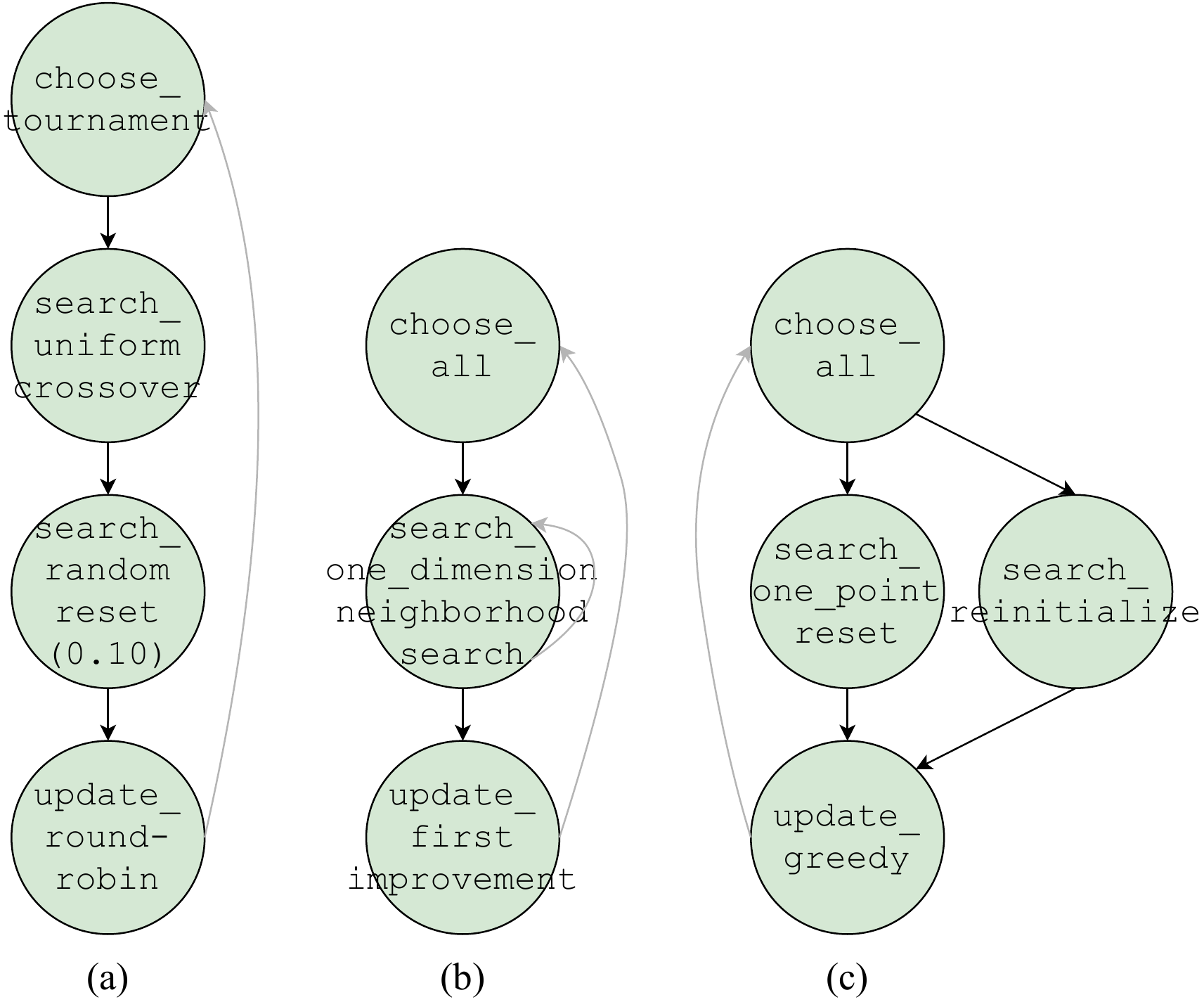}
\caption{Three graph instantiations reflecting different algorithm structures. Algorithms (a) with serial operators, (b) with a recursive operator, (c) with multiple search pathways. Vertices' attribute values are in brackets.}
\label{graph}
\end{figure}

A significant feature of AutoOptLib is that it manipulates directed graph representations of metaheuristic algorithms over the design space. From Figure \ref{graph}, vertices of the graph represent algorithm components; directed edges determine the components' execution flow; attributes of a vertex refer to the component's hyperparameters. Different from other platforms that manipulate the concatenation of identifiers of algorithm components and numerals of hyperparameters (resulting in fixed algorithm structure), the graph manipulated in AutoOptLib is a well-defined format to describe arbitrary orderings of algorithm flow due to three principles: 1) the directed edges enable sequence structures of component compositions; 2) each vertex is open to connecting forward to multiple vertices, allowing branch structures; and 3) cycles realize loop structures. With this flexible representation, AutoOptLib can discover diversified algorithm structures in a single run of design. Users do not need to predefine an algorithm template, but the component composition (vertices) and algorithm structure (graph typology) will be evolved during the design process to fit for solving the target problem.

\subsection{Produce Algorithms} 
In principle, any search method might be employed to search over the design space to produce algorithms. However, as algorithm design is black-box, there is not yet a consensus and dominant method for producing algorithms. The current version of AutoOptLib uses iterative local search \cite{lourencco2003iterated} to manipulate graph typologies (algorithm component compositions) and uses CMA-ES \cite{wierstra2014natural} to optimize vertices' attribute values (components' hyperparameter values), considering the two methods' established track record in algorithm configuration and hyperparameter optimization \cite{hutter2009paramils,hutter2019automated,pushak2022automl}. More methods, e.g., reinforcement learning-based ones, will be added in future versions.

\subsection{Evaluate Performance} 
This module evaluates the produced algorithms' performance according to a design objective. AutoOptLib provides three types of design objectives, i.e., solution quality, runtime, and anytime performance (Table \ref{objective}). AutoOptLib offers four representative evaluation methods, i.e., racing \cite{lopez2016irace}, intensification \cite{hutter2009paramils}, surrogate \cite{lindauer2022smac3}, and exhaustive evaluation (Table \ref{evaluate}). These options enable convenient use for different demands and interests. 

\subsection{Target Problem} 
The target problem acts as external data to support algorithm performance evaluation. For constrained problems, AutoOptLib follows the common practice of the metaheuristic community, i.e., using constraint violations as penalties to discount infeasible solutions. AutoOptLib provides a problem template \textit{prob\_template.m}, by which users can easily implement and interface their problems with the software. An interface with Python problem files is also offered. 

\section{Operating Scheme} \label{sec_operating}
\begin{figure*}[t] 
	\centering
	\includegraphics[width=0.85\linewidth]{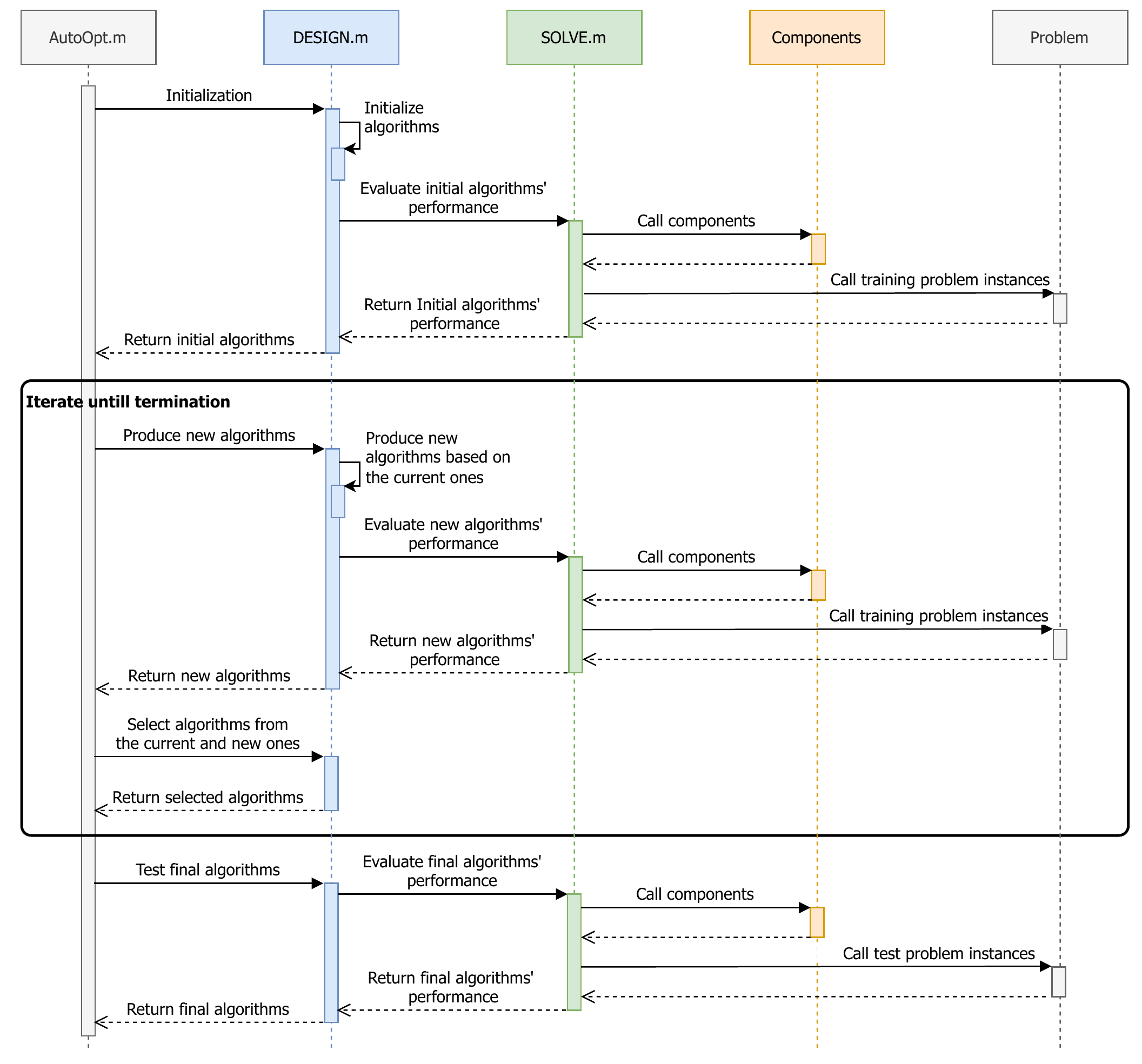}
	\caption{Sequence diagram of AutoOptLib.}
	\label{sequence}
\end{figure*}

The operating scheme of AutoOptLib is depicted in Figure \ref{sequence}. To begin with, the interface function \textit{AutoOpt.m} invokes \textit{DESIGN.m}; \textit{DESIGN.m} uses the \texttt{Initialize()} method to initialize algorithms over the design space. Then, the algorithms' performance on solving the ``training" instances \footnote{Since the distribution of instances of a real problem is often unknown, one has to sample some of the problem instances and target these instances (training instances) during the algorithm design procedure. To avoid the designed algorithms overfit on the training instances, some other instances (test instances) of the target problem are then employed to test the final algorithms after the design procedure terminates.} of the target problem is evaluated by the \texttt{Evaluate()} method. To get the performance, the \texttt{Evaluate()} method invokes the \texttt{SOLVE} class, and \texttt{SOLVE} further calls functions of the algorithms' components and function of the target problem. Finally, the initial algorithms are returned to \textit{AutoOpt.m}.

After initialization, AutoOptLib goes into iterative design. In each iteration, firstly, \textit{AutoOpt.m} invokes \textit{DESIGN.m}; \textit{DESIGN.m} instantiate new algorithms based on the current ones by the \texttt{Disturb()} method. Next, the new algorithms' performance is evaluated in the same scheme as in the initialization. Finally, the \texttt{Select()} method of the \texttt{DESIGN} class is invoked to select promising algorithms from the current and new ones. 

Once the iteration terminates, \textit{AutoOpt.m} invokes the \texttt{Evaluate()} method of the \texttt{DESIGN} class to test the final algorithms' performance on the test instances of the target problem. Then, the final algorithms are returned in \textit{AutoOpt.m}.

The operating scheme has some significant advantages:
\begin{enumerate}
    \item \textit{Metaheuristic component independence.} Functions of algorithm components do not interact with each other but are invoked independently by the \texttt{SOLVE} class. This independence provides flexibility in designing various algorithms and extensibility to new components.
    \item \textit{Design technique packaging.} The design techniques are packaged in different methods (e.g., \texttt{Evaluate()}, \texttt{Estimate()}) of the \texttt{DESIGN} class. Such packaging brings understandability and openness to new techniques without modifying the overall architecture.
    \item \textit{Target problem separation.} The target problem is enclosed separately and does not directly interact with algorithm components and design techniques. This separation allows users to easily interface their problems with AutoOptLib and use it without much knowledge of metaheuristics and design techniques, ensuring AutoOptLib's accessibility to researchers and practitioners from different communities.
\end{enumerate} 

\section{Running} \label{sec_run}
\begin{figure*}[htb] 
	\centering
	\fbox{\includegraphics[width=0.8\linewidth]{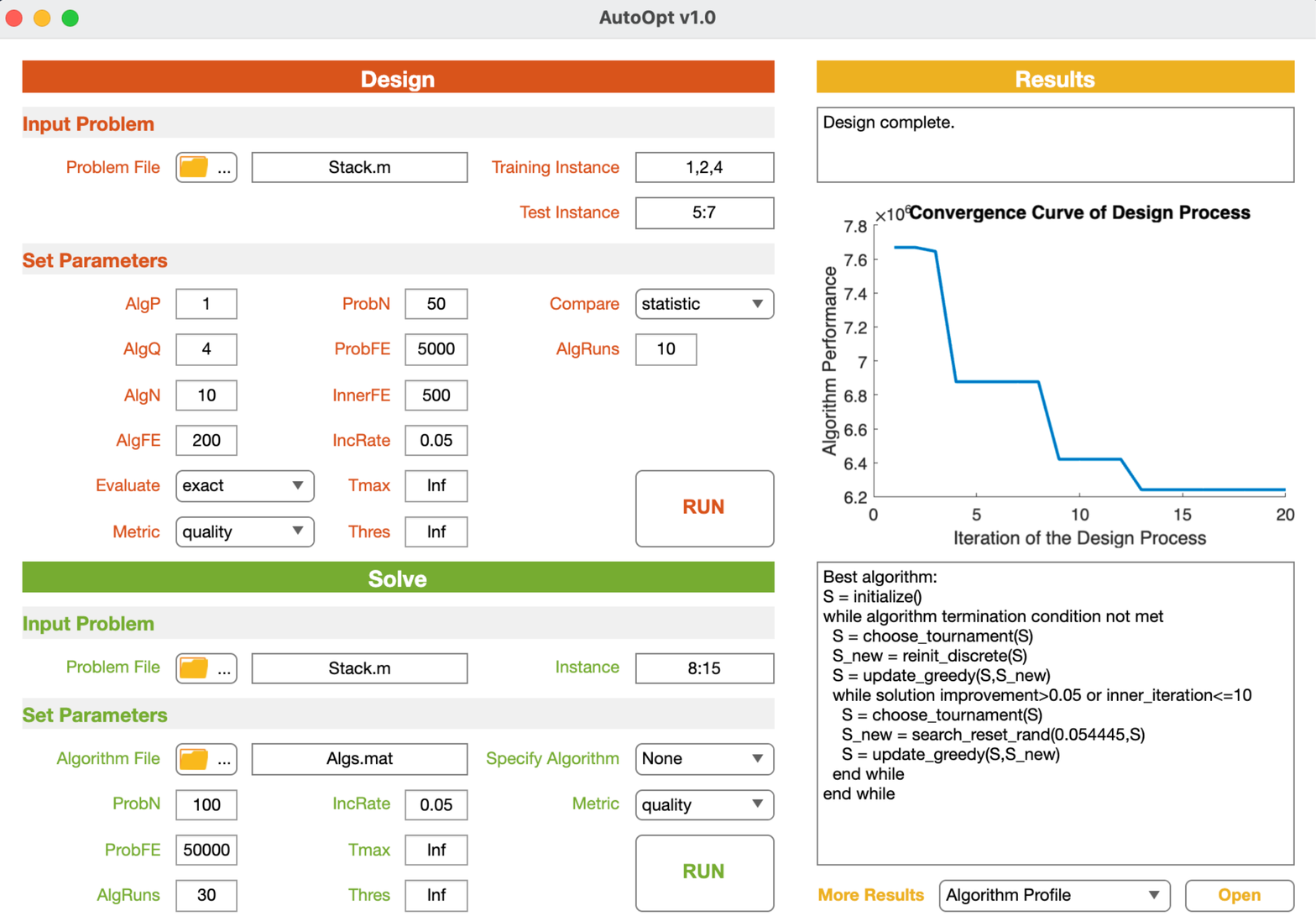}}
	\caption{An example run through the GUI of AutoOptLib.}
	\label{gui}
\end{figure*}

There are two ways to run AutoOptLib. One is by Matlab/Octave command \texttt{AutoOpt(`name1', value1, `name2', value2, ...)}, where name and value refer to the input parameter's name and value, respectively. The parameters specify the target problem, design objective, performance evaluation method, etc.; details are given in the documentation. Another way of running is by the GUI, as shown in Figure \ref{gui}. The GUI manages the whole process, i.e., inputting parameters, monitoring the design process, and visualizing the designed algorithms by simple one-click operations. In detail, the Design panel on the GUI is for designing algorithms for a target problem. All parameters have default values. A description of the parameters is given in the documentation. 

When the running starts, warnings and corresponding solutions to incorrect uses (if any) will be displayed in the text area at the top of the Results panel. The real-time state and progress of the run will also be shown in the area. Once the run terminates, the design process's convergence curve and the best algorithm's pseudocode will be shown in the axes and text areas of the Results panel, respectively. More results can be exported by the pop-up menu at the bottom.

The Solve panel is for solving the target problem by an algorithm. Users can load an algorithm designed by AutoOptLib in the Algorithm File field to solve the target problem. Alternatively, users can choose a baseline algorithm through the pop-up menu of the Specify Algorithm field. We now provide 17 representative metaheuristic algorithms in the menu.

\section{Other Uses} \label{sec_other}
Beyond the primary use of automatically designing algorithms, AutoOptLib provides three important additional functionalities. The first is hyperparameter configuration. Hyperparameter configuration is equivalent to designing an algorithm with predefined component composition but unknown hyperparameter values. Thus, it is easy to perform hyperparameter configuration in AutoOptLib by fixing the component composition and leaving the hyperparameters tunable in the design space. 

The second additional functionality is parameter importance analysis. This can be done by first leaving only one tunable parameter in the design space. Then, AutoOptLib runs and returns the algorithms found during the design process. These algorithms only differ in the values of the tunable parameter. By leaving different parameters tunable in different AutoOptLib runs, users can collect and compare the trends of each parameter's changes versus algorithm performance, subsequently getting insight into each parameter's importance to the algorithm performance.

The last is benchmark comparison. Different design objectives (Table \ref{objective}) and techniques (Table \ref{evaluate}) in AutoOptLib are implemented with the same architecture. This ensures fair comparisons among the objectives or techniques. AutoOptLib can also be used to benchmark comparisons among different algorithms. Users can set AutoOptLib to design multiple algorithms in a single run. These algorithms are built and evaluated in a uniform manner during the design process. Such uniformity ensures a fair comparison among the algorithms.

\section{Experiment} \label{sec_application}
We present two applications of AutoOptLib. One is an academic application to the communication community, by which AutoOptLib illustrates its significance to researchers who seek to use heuristics but are struggling to distinguish the suitability among various algorithms. Another one is an industrial application to supply chain management, by which AutoOptLib demonstrates its practicality to decision-makers without the expertise of equipping a solver for their problems. 
 
\subsection{Academic Application to Passive Beamforming for RIS Aided Communications}
\subsubsection{Problem Description} \label{beamform_problem}
Reconfigurable intelligent surface (RIS) has emerged as an innovative paradigm to achieve a smart and reconfigurable wireless environment for 6G networks \cite{mishra2019channel,yuan2020intelligent}. Generally speaking, it is a planar surface composed of a number of passive reflecting elements, each of which is able to induce a phase shift change on the incident signal. By deploying the RIS and elaborately coordinating the phase shifts, the wireless signals can be flexibly reconfigured to achieve desired directions, in turn improving the wireless communication capacity and reliability. 

\begin{figure}[t]
	\centering
	\includegraphics[width=0.8\linewidth]{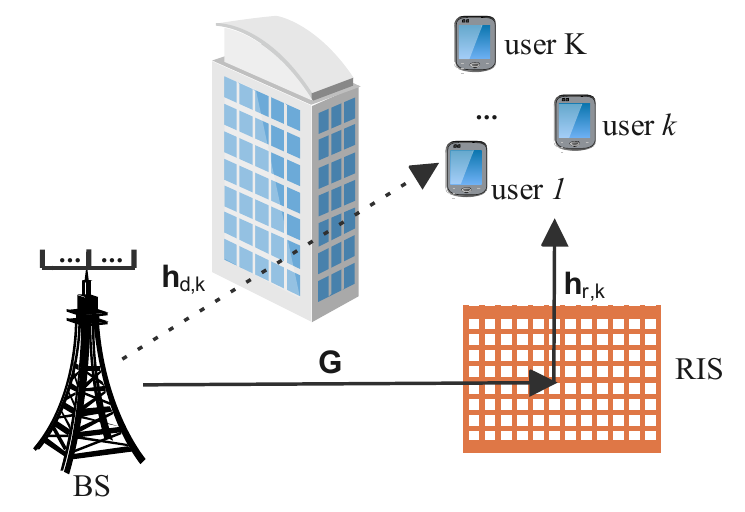}
	\caption{ \scriptsize RIS aided multi-input multi-output downlink system.}
	\label{beanforming}
\end{figure}

We consider a RIS-aided downlink multi-user multiple-input single-output system, as shown in Figure \ref{beanforming}. The base station (BS) with $M$ antennas transmits signals to $K$ single-antenna users. Since some obstacles block the line-of-sight link, a RIS with $N$ elements is deployed to provide non-line-of-sight links. We aim to jointly optimize the continuous active beamforming of BS and discrete phase shifts of RIS for sum-rate maximization, subject to the transmit power constraint. The sum-rate maximization problem is expressed as \cite{yan2022fitness}:
\begin{subequations}
	\label{eq-maxSumRate}
	\begin{align}
        &\max_{\wt_k, \mTa}\quad \sum_{k=1}^{K}\log _{2}(1+\frac{|(\bht_{\fd,k}^\mathrm{H}+\bht_{\fr,k}^\mathrm{H}\mTa\Gt)\wt_k|^2}{\sum_{j\neq k}^K| (\bht_{\fd,k}^\mathrm{H}+\bht_{\fr,k}^\mathrm{H}\mTa\Gt)\wt_{j}|^{2}+\sigma ^{2}}),\tag{\ref{eq-maxSumRate}{a}}\label{eq-Pa} \\ 
        &\quad s.t.\quad \theta_n=\beta_ne^{j\phi_n},\tag{\ref{eq-maxSumRate}{b}}\label{eq-Pb} \\ 
        &\qquad\quad\ \phi_n=\frac{\tau_n2\pi}{2^b}, \tau_n\in\{0,...,2^b-1\},\tag{\ref{eq-maxSumRate}{c}}\label{eq-Pc} \\
        &\qquad\quad\ \sum_{k=1}^{K}\|\mathbf{w}_k\|^{2}\leq P_{T},\tag{\ref{eq-maxSumRate}{d}}\label{eq-Pd}		
	\end{align}
\end{subequations}
where $\wt_k\in\bbC^{M\times 1}$ indicates the active beamforming at BS towards user $k$; $\mTa=diag(\theta_1, ..., \theta_n, ..., \theta_N)$ is a diagonal matrix with RIS phase-shifts as the diagonal values; $\bht_{\fd,k}\in\bbC^{M\times1}$, $\Gt\in\bbC^{N\times M}$, and $\bht_{\fr,k}\in\bbC^{N\times1}$ are the channels BS-user $k$, BS-RIS, and RIS-user $k$; $\beta_n=1$ stands for the reflection coefficients of RIS element $n$; and \eqref{eq-Pd} means the maximum transmit power is limited by $P_T$. 

This problem is a non-convex mixed integer problem, thus is NP-hard. Due to its severe unstructured and rugged landscape, mathematical solvers that decouple RIS elements and estimate each phase shift separately have been demonstrated to be ineligible \cite{yan2022fitness}. Metaheuristics are promising for unstructured, rugged, and highly coupled problems due to their global search ability. Therefore, it is expected that researchers in the communication community benefit from AutoOptLib to quickly obtain an efficient metaheuristic solver.

\subsubsection{AutoOptLib For Beamforming} \label{beamform_experiment}
We utilized AutoOptLib's default design space to design algorithms for the problem. The solution's fitness to the target problem was set as the design objective. Other settings were kept default. We generated 10 problem instances that differ in the number of RIS elements. Five of them were randomly chosen as training instances; the other five were for experimental comparison. 

After running the Design mode of AutoOptLib, we got the best algorithm (termed Alg$^*_{\rm beamform}$) verified by the test instances. Its pseudocode is shown in Algorithm \ref{alg_beamform}. Interestingly, the niching mechanism \texttt{choose\_nich} is involved, which restricts the following uniform crossover (\texttt{cross\_point\_uniform} with crossover rate of $0.1229$) to be performed between solutions within a niching area. The reset operation (\texttt{search\_reset\_one} that resets one entity of the solution) further exploits the niching area. Finally, the round-robin selection (\texttt{update\_round\_robin}) maintains diversity by probably selecting inferior solutions. All these designs indicate that maintaining solution diversity may be necessary for escaping local optima and exploring the unstructured and rugged landscape.  
\begin{algorithm}[t]
	\caption{Pseudocode of Alg$^*_{\rm beamform}$} 
        \scriptsize
	\begin{algorithmic}[1]
		\State $S$ = \texttt{initialize}() \qquad\qquad // initialize solution set $S$
		\While {stopping criterion not met}		
		\State $S$ = \texttt{choose\_nich}($S$)			
		\State $S_{new}$ = \texttt{cross\_point\_uniform}($0.1229,S$)			
		\State $S_{new}$ = \texttt{search\_reset\_one}($S_{new}$)
		\State $S$ = \texttt{update\_round\_robin}($S,S_{new}$)		
		\EndWhile			
	\end{algorithmic}
	\label{alg_beamform}
\end{algorithm}

To investigate the designed algorithm's efficiency for the passive beamforming optimization, we compared it with random beamforming, the representative sequential beamforming \cite{di2020hybrid}\footnote{Sequential beamforming refers to exhaustively enumerating the phase shift of each element one-by-one on the basis of random initial RIS phase shifts.}, and three classic metaheuristic solvers, i.e., the baseline genetic algorithm (GA), iterative local search (ILS), and simulated annealing (SA)\footnote{The discrete GA is consisted by tournament mating selection, one-point crossover, random mutation, and round-robin environmental selection; the mutation rate is set to $0.2$. The ILS and SA perform neighborhood search by randomly resetting one entity of the solution at each iteration.}. The algorithms were executed by the Solve mode of AutoOptLib on the five instances for experimental comparison. All the metaheuristic algorithms conducted population-based search with a population size of 50 for a fair comparison. All algorithms terminated after 50000 function evaluations. We summarize the algorithms' performance in Table \ref{beamform_result}. The performance is measured by final solutions' fitness (reciprocal of all users' service quality). From Table \ref{beamform_result}, sequential beamforming is inferior to most of the metaheuristic solvers. This result confirms the ineligibility of decoupling RIS elements and the need for global metaheuristic search. Among the metaheuristic solvers, Alg$^*_{\rm beamform}$ outperforms others, especially in instances with large numbers of RIS elements (induce high-dimensional rugged landscape). This performance can be attributed to its diversity maintenance ability. All the above demonstrates AutoOptLib's efficiency in the problem.
\begin{table*}[t]
\centering
\caption{Results on the beamforming problem. Best results are in bold.}
\label{beamform_result}
\begin{threeparttable}
\begin{tabular}{p{2cm}p{2.5cm}<{\centering}p{2.5cm}<{\centering}p{2.5cm}<{\centering}p{2.5cm}<{\centering}p{2.5cm}<{\centering}}
\toprule
\multirow{2}{*}{Algorithm} & \multicolumn{5}{c}{Number of RIS elements in the problem instances}\\
                       & 120 & 160 & 280 & 320 & 400\\
\midrule
Alg$^*_{\rm beamform}$ & \textbf{0.0332}$\pm$5.05E-04 & \textbf{0.0312}$\pm$4.84E-04 & \textbf{0.0281}$\pm$1.57E-04 & \textbf{0.0272}$\pm$6.76E-04 & \textbf{0.0260}$\pm$1.11E-04 \\
Random                 & 0.0442$\pm$7.94E-04 & 0.0425$\pm$6.56E-04 & 0.0402$\pm$8.30E-04 & 0.0390$\pm$6.67E-04 & 0.0375$\pm$1.88E-04 \\
Sequential             & 0.0382$\pm$6.19E-04 & 0.0387$\pm$6.75E-04 & 0.0374$\pm$4.17E-04 & 0.0369$\pm$4.38E-04 & 0.0354$\pm$8.27E-04 \\
GA                     & 0.0369$\pm$3.30E-04 & 0.0356$\pm$1.00E-04 & 0.0337$\pm$4.26E-04 & 0.0333$\pm$1.04E-04 & 0.0322$\pm$6.96E-04 \\
ILS                    & 0.0333$\pm$3.74E-04 & 0.0314$\pm$2.49E-04 & 0.0285$\pm$1.19E-04 & 0.0279$\pm$1.82E-04 & 0.0278$\pm$1.15E-04 \\
SA                     & 0.0398$\pm$5.59E-04 & 0.0388$\pm$7.75E-04 & 0.0369$\pm$3.27E-04 & 0.0360$\pm$4.18E-04 & 0.0355$\pm$9.50E-04 \\
\bottomrule
\end{tabular}
\begin{tablenotes}
    \item Alg$^*_{\rm beamform}$ is the algorithm designed by AutoOptLib.
\end{tablenotes}
\end{threeparttable}
\end{table*}

\subsection{Industrial Application to Raw Material Stacking in Supply Chain Management}
\subsubsection{Problem Description}
\begin{figure}[t]
	\centering
	\includegraphics[width=0.9\linewidth]{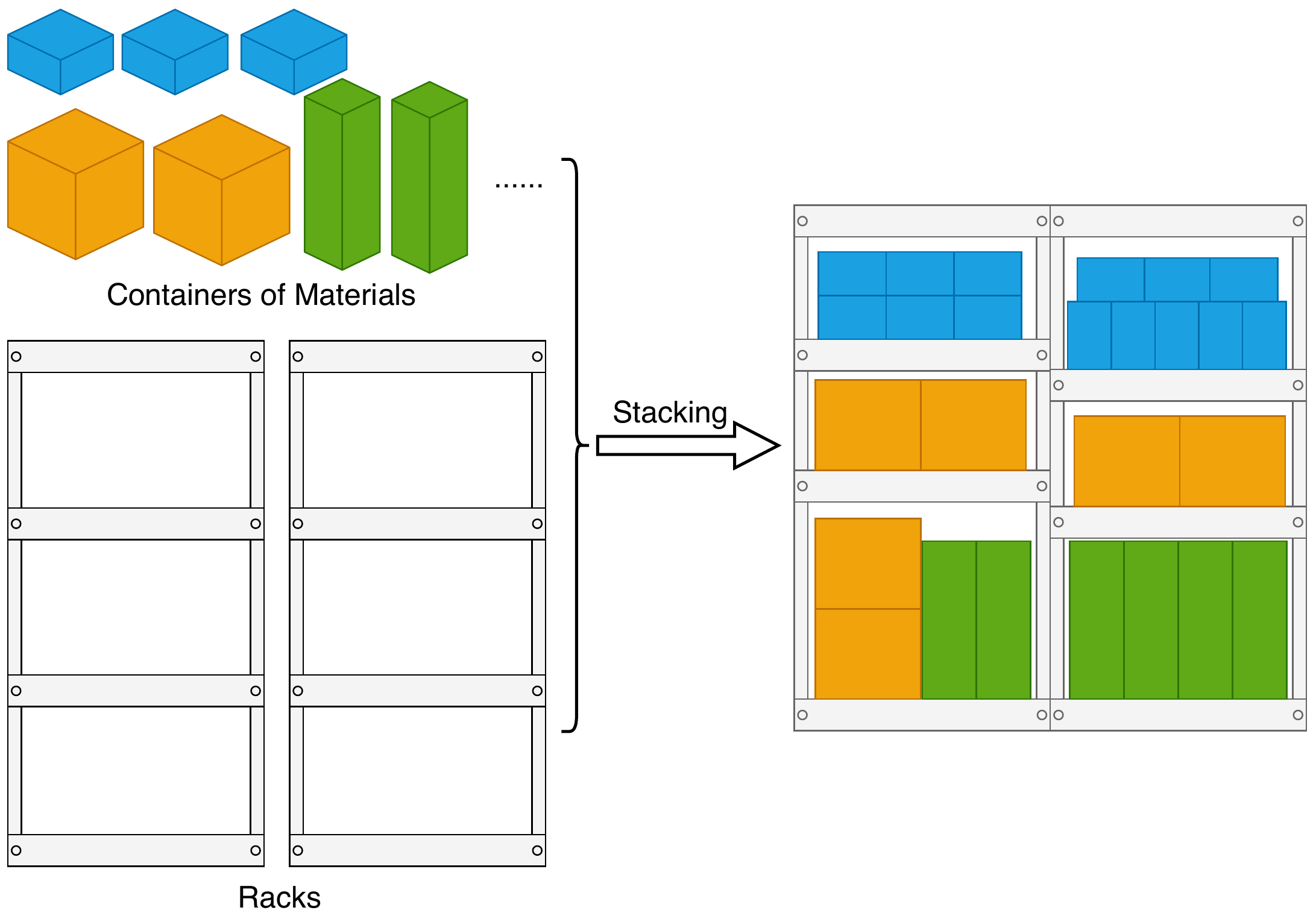}
	\caption{\scriptsize Graphical illustration of material stacking.}
	\label{stack}
\end{figure}

The problem comes from the supply chain department of a biomedical electronics company. Raw material stacking is one of the main tasks in the supply chain management of the company. This task refers to 1) shaping the racks in the warehouse to fit for stacking the containers of raw materials, and 2) placing the containers in appropriate locations on the racks. The stacking problem consists of these two parts, i.e., shaping racks and placing materials' containers. The stacking should be effective in space consumption and convenient for picking up and sending materials to the downstream manufacturing department. A graphical illustration of the problem is shown in Figure \ref{stack}.

Specifically, decision variables of the problem are the heights of each floor of the racks (racks are with fixed body sizes but discretely adjustable floor heights) and materials' ways of stacking (discrete locations and orientations for placing each material's container on a certain floor of a rack). The objective of the problem is to minimize the space consumption of racks in stacking the materials and, at the same time, to satisfy the following rules: heavy materials stack on lower floors of the racks; materials of the same kind stack together; unique materials for a product stack near each other; common materials that serve multiple products stack at a place near all the served products; and materials with higher picking up frequencies stack closer to the shipping area. The problem has 39 instances that differ in the input data (materials and racks to be considered).

The problem is challenging in the following aspects: 1) the stacking problem has been proven to be NP-hard \cite{liu2019product}; 2) materials for the precise biomedical electronic products, i.e., computed tomography scanners and blood detection machines, are large-scale and heterogeneous (up to 4000 kinds of materials with a total amount of over 50000 units in the target problem); and 3) the problem is heavily constrained due to the complicated rules to be satisfied. These challenges make mathematical solvers impractical on the problem. Metaheuristic solvers could meet these challenges, but practitioners are short of expertise to manage an efficient use of the solvers. Leveraging automated techniques by AutoOptLib would be desirable in this regard.

\subsubsection{AutoOptLib for Material Stacking} \label{stack_experiment}
We employed AutoOptLib's default design space to design algorithms for the stacking problem. The solution's fitness to the target problem was set as the design objective, and racing was used as the algorithm evaluation method. 20 instances were randomly chosen from the total 39 instances; 10 of the chosen instances were employed as training instances, while the other 10 were for comparing the designed algorithm with baselines. Other settings of AutoOptLib were kept default. 

After running the Design mode of AutoOptLib, we got the best algorithm (termed Alg$^*_{\rm stack}$) with pseudocode shown in Algorithm \ref{alg_stack}. Alg$^*_{\rm stack}$ is a GA-style algorithm, in which the one-point crossover and random reset are executed sequentially to search solutions, and the reset probability is tuned as $0.1342$. This algorithm indicates that for the large-scale discrete problem, a large degree of solution recombination (by one-point crossover) and exploration (reset each entity of the solution with a probability of $0.1342$) may be fit for keeping feasibility regarding the constraints and ensuring search efficiency over the large-scale solution space.  
\begin{algorithm}[t]
	\caption{Pseudocode of Alg$^*_{\rm stack}$} 
        \scriptsize
	\begin{algorithmic}[1]
		\State $S$ = \texttt{initialize}() \qquad\qquad // initialize solution set $S$
		\While {stopping criterion not met}		
		\State $S$ = \texttt{choose\_roulette\_wheel}($S$)			
		\State $S_{new}$ = \texttt{cross\_point\_one}($S$)			
		\State $S_{new}$ = \texttt{search\_reset\_rand}($0.1342,S_{new}$)
		\State $S$ = \texttt{update\_round\_robin}($S,S_{new}$)		
		\EndWhile			
	\end{algorithmic}
	\label{alg_stack}
\end{algorithm}

We further compared Alg$^*_{\rm stack}$ with the baseline random search (RS) and three classic metaheuristic solvers, i.e., discrete GA, ILS, and SA. The discrete GA, ILS, and SA are the same as those employed in Section \ref{beamform_experiment}. All runs were with the same settings and termination condition defaulted in AutoOptLib. We report the algorithms' performance over the 10 instances in Table \ref{stack_result}. The performance is measured by final solutions' fitness. According to Table \ref{stack_result}, Alg$^*_{\rm stack}$ obtains much better performance than the compared algorithms, demonstrating AutoOptLib's efficiency on the problem. Alg$^*_{\rm stack}$’s solutions had been presented to the material stacking practitioners. Solutions to some instances (for certain materials and racks) had been verified in practice and reported to reduce the space waste rate to lower than 20\%, which was considered satisfactory.
\begin{table}[t]
\centering
\caption{Results on the stacking problem. Best results are in bold.}
\label{stack_result}
\begin{tabular}{p{1.5cm}p{3cm}<{\centering}p{3cm}<{\centering}}
\toprule
Algorithm & Mean & Std. \\
\midrule
Alg$^*_{\rm stack}$ & \textbf{31914538.82} & 6.88E+05 \\
GA & 33024530.26 & 6.77E+05 \\
ILS & 32868538.15 & 5.65E+05 \\
SA & 37233538.82 & 8.87E+05 \\
RS & 44178551.09 & 4.93E+05 \\
\bottomrule
\end{tabular}
\end{table}

\section{Conclusion} \label{sec_conclusion}
This paper has presented AutoOptLib for accessible automated design of metaheuristic optimizers. AutoOptLib provides throughout support to the design process, including plenty of algorithm components for constructing the design space, flexible representation for discovering diverse algorithm structures, different design objectives and techniques, and GUI. Applications to two real problem-solving scenarios have demonstrated AutoOptLib's efficiency in tailoring satisfactory metaheuristic solvers to complicated problems. In the future, we will continually maintain the platform to incorporate more metaheuristic components, support more types of problems (e.g., mixed-integer and multi-objective ones), add learning-based design methods, and add the function of outputting the source code of the designed optimizers for deploying them on specific devices. 

\bibliographystyle{IEEEtran}
\bibliography{Refs}

\appendix
\setcounter{table}{0}   
\setcounter{figure}{0}
\renewcommand{\thetable}{A\arabic{table}}
\renewcommand{\thefigure}{A\arabic{figure}}

\begin{table*}[htbp]
\centering
\scriptsize
\caption{Metaheuristic algorithm components provided in AutoOptLib.}
\label{component}
\begin{tabular}{p{4cm}p{13cm}}
\toprule
Component & Description \\
\midrule
\textbf{Continuous search:}           & \\
\texttt{cross\_arithmetic}            & Whole arithmetic crossover \\
\texttt{cross\_sim\_binary}           & Simulated binary crossover \\
\texttt{cross\_point\_one}            & One-point crossover \\
\texttt{cross\_point\_two}            & Two-point crossover \\
\texttt{cross\_point\_n}              & \textit{n}-point crossover \\
\texttt{cross\_point\_uniform}        & Uniform crossover \\
\texttt{search\_cma}                  & The evolution strategy with covariance matrix adaption \\
\texttt{search\_eda}                  & The estimation of distribution \\
\texttt{search\_mu\_cauchy}           & Cauchy mutation \\
\texttt{search\_mu\_gaussian}         & Gaussian mutation \\
\texttt{search\_mu\_polynomial}       & Polynomial mutation  \\
\texttt{search\_mu\_uniform}          & Uniform mutation  \\
\texttt{search\_pso}                  & Particle swarm optimization's particle fly and update  \\
\texttt{search\_de\_random}           & The "random/1" differential mutation  \\
\texttt{search\_de\_current}          & The "current/1" differential mutation \\
\texttt{search\_de\_current\_best}    & The "current-to-best/1" differential mutation \\
\texttt{reinit\_continuous}           & Random reinitialization for continuous problems \\
\midrule                             
\textbf{Discrete search:}             &  \\
\texttt{cross\_point\_one}            & One-point crossover \\
\texttt{cross\_point\_two}            & Two-point crossover \\
\texttt{cross\_point\_n}              & \textit{n}-point crossover \\
\texttt{cross\_point\_uniform}        & Uniform crossover \\
\texttt{search\_reset\_one}           & Reset a randomly selected entity to a random value \\
\texttt{search\_reset\_rand}          & Reset each entity to a random value with a probability \\
\texttt{search\_reset\_creep}         & Add a small positive or negative value to each entity with a probability, for problems with ordinal attributes \\
\texttt{reinit\_discrete}             & Random reinitialization for discrete problems \\ 
\midrule
\textbf{Permutation search:}          &  \\
\texttt{cross\_order\_two}            & Two-order crossover \\
\texttt{cross\_order\_n}              & \textit{n}-order crossover \\
\texttt{search\_swap}                 & Swap two randomly selected entities \\
\texttt{search\_swap\_multi}          & Swap each pair of entities between two randomly selected indices \\
\texttt{search\_scramble}             & Scramble all the entities between two randomly selected indices \\
\texttt{search\_insert}               & Randomly select two entities, insert the second entity to the position following the first one \\
\texttt{reinit\_permutation}          & Random reinitialization for permutation problems \\
\midrule
\multicolumn{2}{l}{\textbf{Choose where to search from:}} \\
\texttt{choose\_roulette\_wheel}      & Roulette wheel selection \\
\texttt{choose\_tournament}           & \textit{K}-tournament selection \\
\texttt{choose\_traverse}             & Choose each of the current solutions to search from \\
\texttt{choose\_cluster}              & Brain storm optimization's idea picking up for choosing solutions to search from  \\
\texttt{choose\_nich}                 & Adaptive niching based on the nearest-better clustering \\
\midrule
\multicolumn{2}{l}{\textbf{Select promising solutions:}} \\
\texttt{update\_always}               & Always select new solutions \\
\texttt{update\_greedy}               & Select the best solutions \\
\texttt{update\_pairwise}             & Select the better solution from each pair of old and new solutions \\
\texttt{update\_round\_robin}         & Select solutions by round-robin tournament \\
\texttt{update\_simulated\_annealing} & Simulated annealing's update mechanism, i.e., accept worse solution with a probability \cite{kirkpatrick1983optimization} \\
\midrule
\textbf{Archive:}                     & \\
\texttt{archive\_best}                & Collect the best solutions found so far \\
\texttt{archive\_diversity}           & Collect most diversified solutions found so far \\
\texttt{archive\_tabu}                & The tabu list \\
\bottomrule
\end{tabular}
\end{table*}

\begin{table*}[htbp]
\centering
\caption{Design objectives involved in AutoOptLib.}
\label{objective}
\scriptsize
\begin{tabular}{p{4cm}p{13cm}}
\toprule
Objective & Description \\
\midrule
\texttt{quality}    & Algorithm's solution quality on the target problem within a fixed evoluation budget. \\
\texttt{runtimeFE}  & Algorithm's running time (number of function evaluations) till reaching a performance threshold on the target problem. \\
\texttt{runtimeSec} & Algorithm's running time (wall clock time, in second) till reaching a performance threshold on the target problem. \\
\texttt{auc}        & The area under the curve (AUC) of empirical cumulative distribution function of running time, measuring the anytime performance. \\
\bottomrule
\end{tabular}
\end{table*}

\begin{table*}[htbp]
\centering
\caption{Algorithm performance evaluation methods provided in AutoOptLib.}
\label{evaluate}
\scriptsize
\begin{tabular}{p{4cm}p{13cm}}
\toprule
Method & Description \\
\midrule
\texttt{racing} & Save algorithm evaluations by stopping evaluating on the next instance if performance is statistically worse than at least another algorithm \cite{lopez2016irace} . \\
\texttt{intensification} & Save algorithm evaluations by stopping evaluating on the next instance if performance is worse than the incumbent \cite{hutter2009paramils}. \\
\texttt{approximate}      &  Use low-complexity surrogate to approximate the algorithms' performance without full evaluation. \\
\texttt{exhaustive}            &  Exactly run all the algorithms on all target problem instances. \\
\bottomrule
\end{tabular}
\end{table*}

\end{document}